\documentclass[10pt,twocolumn,letterpaper]{article}

\usepackage{cvpr}
\usepackage{times}
\usepackage{epsfig}
\usepackage{graphicx}
\usepackage{amsmath}
\usepackage{amssymb}

\usepackage{sg-macros}
\usepackage{enumitem}
\usepackage{booktabs}
\usepackage{makecell}

\usepackage[pagebackref=true,breaklinks=true,letterpaper=true,colorlinks,bookmarks=false]{hyperref}

\cvprfinalcopy 

\def\cvprPaperID{1027} 

\ifcvprfinal\fi

\makeatletter
\let\ps@plain\ps@empty
\makeatother

\begin{document}

\title{ Vision-and-Language Navigation: Interpreting visually-grounded\\navigation instructions in real environments }

\author{Peter Anderson\textsuperscript{1} \hspace{20pt} Qi Wu\textsuperscript{2} \hspace{20pt} Damien Teney\textsuperscript{2} \hspace{20pt} Jake Bruce\textsuperscript{3} \hspace{20pt} Mark Johnson\textsuperscript{4}\\ Niko S\"{u}nderhauf\textsuperscript{3} \hspace{20pt} Ian Reid\textsuperscript{2} \hspace{20pt}
	Stephen Gould\textsuperscript{1} \hspace{20pt} Anton van den Hengel\textsuperscript{2}\\
	\normalsize{
		\textsuperscript{1}Australian National University \space \textsuperscript{2}University of Adelaide \space \textsuperscript{3}Queensland University of Technology \space \textsuperscript{4}Macquarie University
	}\\
	\tt\small\textsuperscript{1}firstname.lastname@anu.edu.au, \tt\small\textsuperscript{3}jacob.bruce@hdr.qut.edu.au,
	\tt\small\textsuperscript{3}niko.suenderhauf@qut.edu.au\\
	\tt\small\textsuperscript{2}\{qi.wu01,damien.teney,ian.reid,anton.vandenhengel\}@adelaide.edu.au,
	\tt\small\textsuperscript{4}mark.johnson@mq.edu.au
}

\maketitle

\begin{abstract}
A robot that can carry out a natural-language instruction has been a dream since before the Jetsons cartoon series imagined a life of leisure mediated by a fleet of attentive robot helpers.  It is a dream that remains stubbornly distant. However, recent advances in vision and language methods have made incredible progress in closely related areas. This is significant because a robot interpreting a natural-language navigation instruction on the basis of what it sees is carrying out a vision and language process that is similar to Visual Question Answering.  Both tasks can be interpreted as visually grounded sequence-to-sequence translation problems, and many of the same methods are applicable. To enable and encourage the application of vision and language methods to the problem of interpreting visually-grounded navigation instructions, we present the Matterport3D Simulator -- a large-scale reinforcement learning environment based on real imagery~\cite{Matterport3D}. Using this simulator, which can in future support a range of embodied vision and language tasks, we provide the first benchmark dataset for visually-grounded natural language navigation in real buildings -- the Room-to-Room (R2R) dataset\footnote{https://bringmeaspoon.org}. 	
\end{abstract}

\section{Introduction}

\begin{figure}[htb]
	\begin{center}
		\includegraphics[width=1\linewidth]{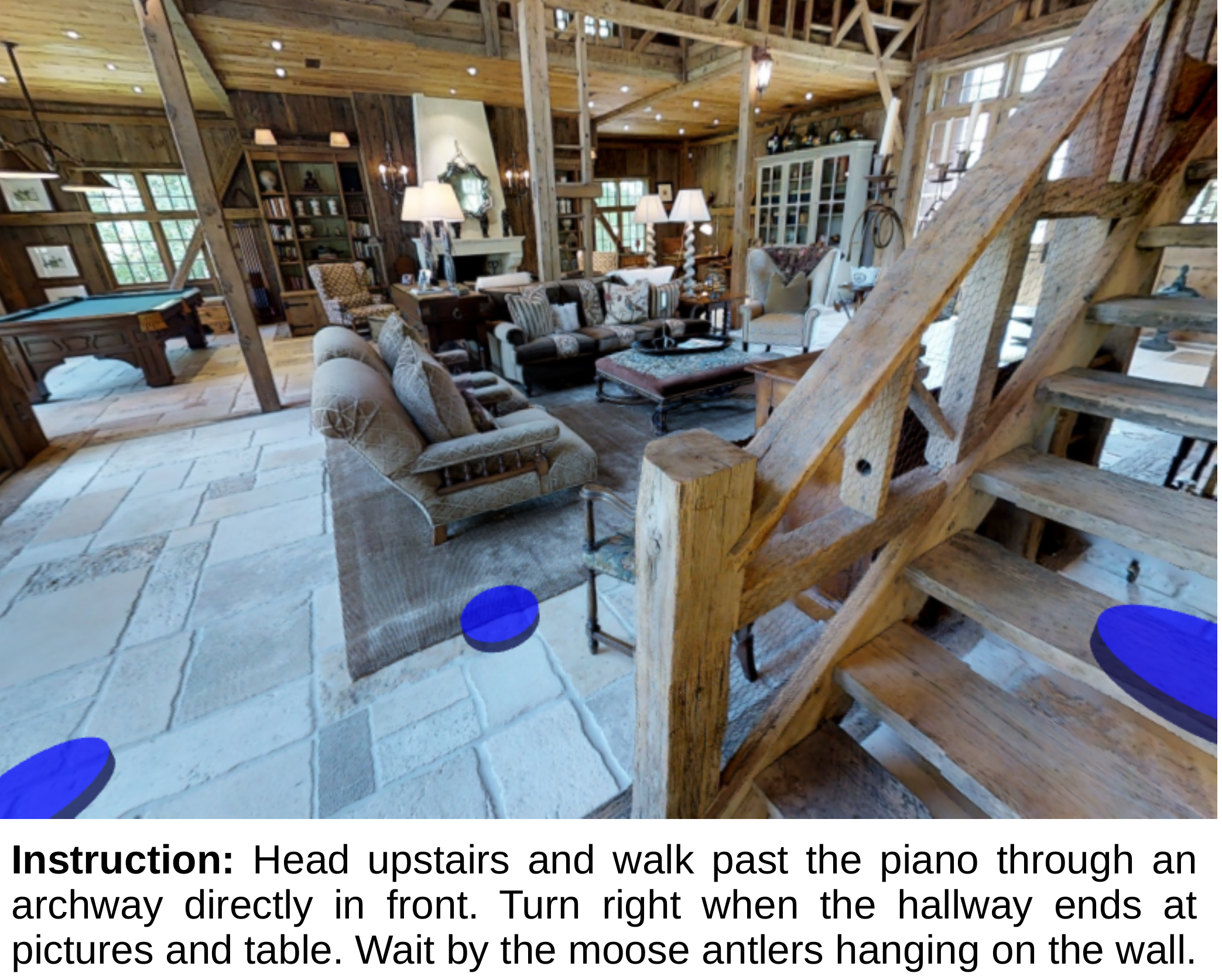}
	\end{center}
	\caption{	
		Room-to-Room (R2R) navigation task. We focus on executing natural language navigation instructions in previously unseen real-world buildings. The agent's camera can be rotated freely. Blue discs indicate nearby (discretized) navigation options.}
	\label{fig:concept}
\end{figure}

The idea that we might be able to give general, verbal instructions to a robot and have at least a reasonable probability that it will carry out the required task is one of the long-held goals of robotics, and artificial intelligence (AI). Despite significant progress, there are a number of major technical challenges that need to be overcome before robots will be able to perform general tasks in the real world. One of the primary requirements will be new techniques for linking natural language to vision and action in \textit{unstructured, previously unseen environments}. It is the navigation version of this challenge that we refer to as Vision-and-Language Navigation (VLN).  

Although interpreting natural-language navigation instructions has received significant attention previously~\cite{chaplot2017gated,chen2011learning,guadarrama2013grounding,mei2016listen,misra2017mapping,tellex2011understanding}, it is the recent success of recurrent neural network methods for the joint interpretation of images and natural language that motivates the VLN task, and the associated Room-to-Room (R2R) dataset described below.  The dataset particularly has been designed to simplify the application of vision and language methods to what might otherwise seem a distant problem.

\begin{figure*}[t]
	\begin{center}
		\includegraphics[width=0.75\linewidth]{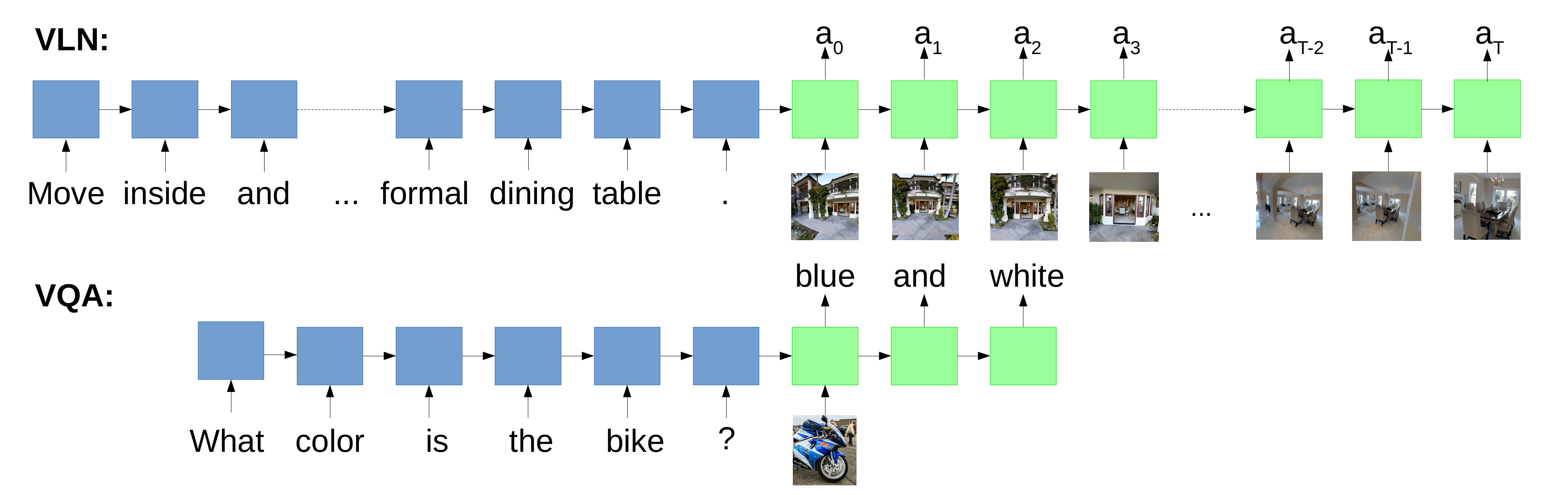}
	\end{center}
	\caption{Differences between Vision-and-Language Navigation (VLN)  and Visual Question Answering (VQA). Both tasks can be formulated as visually grounded sequence-to-sequence transcoding problems. However, VLN sequences are much longer and, uniquely among vision and language benchmark tasks using real images, the model outputs actions $\langle a_0, a_1, \dots a_T \rangle$ that manipulate the camera viewpoint.}
	\label{fig:vqa}
\end{figure*}

Previous approaches to natural language command of robots have often neglected the visual information processing aspect of the problem. Using rendered, rather than real images~\cite{beattie2016deepmind,Kempka2016ViZDoom,zhu2017icra}, for example, constrains the set of visible objects to the set of hand-crafted models available to the renderer.  This turns the robot's challenging open-set problem of relating real language to real imagery into a far simpler closed-set classification problem.  The natural extension of this process is that adopted in works where the images are replaced by a set of labels~\cite{chen2011learning,tellex2011understanding}.  Limiting the variation in the imagery inevitably limits the variation in the navigation instructions also. What distinguishes the VLN challenge is that the agent is required to interpret a previously \textit{unseen} natural-language navigation command in light of images generated by a previously \textit{unseen} real environment.  The task thus more closely models the distinctly open-set nature of the underlying problem.

To enable the reproducible evaluation of VLN methods, we present the Matterport3D Simulator. The simulator is a large-scale interactive reinforcement learning (RL) environment constructed from the Matterport3D dataset~\cite{Matterport3D} which contains 10,800 densely-sampled panoramic RGB-D images of 90 real-world building-scale indoor environments. Compared to synthetic RL environments~\cite{beattie2016deepmind,Kempka2016ViZDoom, zhu2017icra}, the use of real-world image data preserves visual and linguistic richness, maximizing the potential for trained agents to be transferred to real-world applications. 

Based on the Matterport3D environments, we collect the Room-to-Room (R2R) dataset containing 21,567 open-vocabulary, crowd-sourced navigation instructions with an average length of 29 words. Each instruction describes a trajectory traversing typically multiple rooms. As illustrated in \figref{fig:concept}, the associated task requires an agent to follow natural-language instructions to navigate to a goal location in a previously unseen building. We investigate the difficulty of this task, and particularly the difficulty of operating in unseen environments, using several baselines and a sequence-to-sequence model based on methods successfully applied to other vision and language tasks~\cite{VQA,Chen2015,balanced_vqa_v2}.

In summary, our main contributions are:
\setlist{nolistsep}
\begin{enumerate}[noitemsep]
	\item We introduce the Matterport3D Simulator, a software framework for visual reinforcement learning using the Matterport3D panoramic RGB-D dataset~\cite{Matterport3D};
	\item We present Room-to-Room (R2R), the first benchmark dataset for Vision-and-Language Navigation in real, previously unseen, building-scale 3D environments;
	\item We apply sequence-to-sequence neural networks to the R2R dataset, establishing several baselines. 
\end{enumerate}

\noindent
The simulator, R2R dataset and baseline models are available through the project website at https://bringmeaspoon.org.
\section{Related Work}

\paragraph{Navigation and language}
Natural language command of robots in unstructured environments has been a research goal for several decades~\cite{winograd1971procedures}. However, many existing approaches abstract away the problem of visual perception to some degree. This is typically achieved either by assuming that the set of all navigation goals, or objects to be acted upon, has been enumerated, and that each will be identified by label~\cite{chen2011learning, tellex2011understanding}, or by operating in visually restricted environments requiring limited perception~\cite{chaplot2017gated,guadarrama2013grounding,huang2010natural,kollar2010toward,macmahon2006walk,mei2016listen,vogel2010learning}. Our work contributes for the first time a navigation benchmark dataset that is both linguistically and visually rich, moving closer to real scenarios while still enabling reproducible evaluations.  

\vspace{-0.3cm}
\paragraph{Vision and language}
The development of new benchmark datasets for image captioning~\cite{Chen2015}, visual question answering (VQA)~\cite{VQA,balanced_vqa_v2} and visual dialog~\cite{visualdialog} has spurred considerable progress in vision and language understanding, enabling models to be trained end-to-end on raw pixel data from large datasets of natural images. However, although many tasks combining visual and linguistic reasoning have been motivated by their potential robotic applications~\cite{VQA, visualdialog,KazemzadehOrdonezMattenBergEMNLP14,mao2015generation,MovieQA}, none of these tasks allow an agent to move or control the camera. As illustrated in \figref{fig:vqa}, our proposed R2R benchmark addresses this limitation, which also motivates several concurrent works on embodied question answering~\cite{embodiedqa, iqa}. 

\vspace{-0.3cm}
\paragraph{Navigation based simulators}
Our simulator is related to existing 3D RL environments based on game engines, such as ViZDoom~\cite{Kempka2016ViZDoom}, DeepMind Lab~\cite{beattie2016deepmind} and AI2-THOR~\cite{thor}, as well as a number of newer environments developed concurrently including HoME~\cite{brodeur2017home}, House3D~\cite{house3d}, MINOS~\cite{savva2017minos}, CHALET~\cite{chalet} and Gibson Env~\cite{zamir2018embodied}. The main advantage of our framework over synthetic environments~\cite{thor,brodeur2017home,house3d,chalet} is that all pixel observations come from natural images of real scenes, ensuring that almost every coffee mug, pot-plant and wallpaper texture is unique. This visual diversity and richness is hard to replicate using a limited set of 3D assets and textures. Compared to MINOS~\cite{savva2017minos}, which is also based on Matterport data~\cite{Matterport3D}, we render from panoramic images rather than textured meshes. Since the meshes have missing geometry -- particularly for windows and mirrors -- our approach improves visual realism but limits navigation to discrete locations (refer \secref{sec:simsub} for details). Our approach is similar to the (much smaller) Active Vision Dataset~\cite{ammirato2017dataset}.

\vspace{-0.3cm}
\paragraph{RL in navigation}
A number of recent papers use reinforcement learning (RL) to train navigational agents~\cite{kulkarni2016deep,tai2016towards,tessler2017deep,zhu2017icra,gupta2017cognitive}, although these works do not address language instruction. The use of RL for language-based navigation has been studied in \cite{chaplot2017gated} and \cite {misra2017mapping}, however, the settings are visually and linguistically less complex. For example, Chaplot \etal \cite{chaplot2017gated} develop an RL model to execute template-based instructions in Doom environments~\cite{Kempka2016ViZDoom}. Misra \etal \cite {misra2017mapping} study complex language instructions in a fully-observable blocks world. By releasing our simulator and dataset, we hope to encourage further research in more realistic partially-observable settings.

\section{Matterport3D Simulator}
\label{sec:sim}

In this section we introduce the Matterport3D Simulator, a new large-scale visual reinforcement learning (RL) simulation environment for the research and development of intelligent agents based on the Matterport3D dataset~\cite{Matterport3D}. The Room-to-Room (R2R) navigation dataset is discussed in \secref{sec:dataset}.

\subsection{Matterport3D Dataset}

Most RGB-D datasets are derived from video sequences; e.g. NYUv2~\cite{NYUv2}, SUN RGB-D~\cite{SUNrgbd} and ScanNet~\cite{dai2017scannet}. These datasets typically offer only one or two paths through a scene, making them inadequate for simulating robot motion. In contrast to these datasets, the recently released Matterport3D dataset~\cite{Matterport3D} contains a comprehensive set of panoramic views. To the best of our knowledge it is also the largest currently available RGB-D research dataset.

In detail, the Matterport3D dataset consists of 10,800 panoramic views constructed from 194,400 RGB-D images of 90 building-scale scenes. On average, panoramic viewpoints are distributed throughout the entire walkable floor plan of each scene at an average separation of 2.25m. Each panoramic view is comprised of 18 RGB-D images captured from a single 3D position at the approximate height of a standing person. Each image is annotated with an accurate 6 DoF camera pose, and collectively the images capture the entire sphere except the poles. The dataset also includes globally-aligned, textured 3D meshes annotated with class and instance segmentations of regions (rooms) and objects. 

In terms of visual diversity, the selected Matterport scenes encompass a range of buildings including houses, apartments, hotels, offices and churches of varying size and complexity. These buildings contain enormous visual diversity, posing real challenges to computer vision. Many of the scenes in the dataset can be viewed in the Matterport 3D spaces gallery\footnote{https://matterport.com/gallery/}. 

\subsection{Simulator}
\label{sec:simsub}

\subsubsection{Observations}
\label{sec:obs}
To construct the simulator, we allow an embodied agent to virtually `move' throughout a scene by adopting poses coinciding with panoramic viewpoints. Agent poses are defined in terms of 3D position $v \in V$, heading $\psi \in [0, 2\pi)$, and camera elevation $\theta \in [-\frac{\pi}{2}, \frac{\pi}{2}]$, where $V$ is the set of 3D points associated with panoramic viewpoints in the scene. At each step $t$, the simulator outputs an RGB image observation $o_t$ corresponding to the agent's first person camera view. Images are generated from perspective projections of precomputed cube-mapped images at each viewpoint. Future extensions to the simulator will also support depth image observations (RGB-D), and additional instrumentation in the form of rendered object class and object instance segmentations (based on the underlying Matterport3D mesh annotations). 
\subsubsection{Action Space}
\label{sec:action-space}
The main challenge in implementing the simulator is determining the state-dependent action space. Naturally, we wish to prevent agents from teleporting through walls and floors, or traversing other non-navigable regions of space. Therefore, at each step $t$ the simulator also outputs a set of next step reachable viewpoints $W_{t+1} \subseteq V$. Agents interact with the simulator by selecting a new viewpoint $v_{t+1} \in W_{t+1}$, and nominating camera heading ($\Delta \psi_{t+1}$) and elevation ($\Delta \theta_{t+1}$) adjustments. Actions are deterministic.

To determine $W_{t+1}$, for each scene the simulator includes a weighted, undirected graph over panoramic viewpoints, $G = \langle V,E \rangle$, such that the presence of an edge signifies a robot-navigable transition between two viewpoints, and the weight of that edge reflects the straight-line distance between them. To construct the graphs, we ray-traced between viewpoints in the Matterport3D scene meshes to detect intervening obstacles. To ensure that motion remains localized, we then removed edges longer than 5m. Finally, we manually verified each navigation graph to correct for missing obstacles not captured in the meshes (such as windows and mirrors). 

Given navigation graph $G$, the set of next-step reachable viewpoints is given by: 
\vspace{-2pt}
\begin{align}
\label{eq:reachable}
W_{t+1} = \big\{ v_t \big\} \cup \big\{v_i \in V \mid \langle v_{t},v_i \rangle \in E \land v_i \in P_t \big\}
\end{align}
\noindent
where $v_t$ is the current viewpoint, and $P_t$ is the region of space enclosed by the left and right extents of the camera view frustum at step $t$. In effect, the agent is permitted to follow any edges in the navigation graph, provided that the destination is within the current field of view, or visible by glancing up or down\footnote{This avoids forcing the agent to look at the floor every time it takes a small step.}. Alternatively, the agent always has the choice to remain at the same viewpoint and simply move the camera. 

\begin{figure}[t]
	\begin{center}
		\includegraphics[trim={0 0 0cm 0},clip,width=1\linewidth]{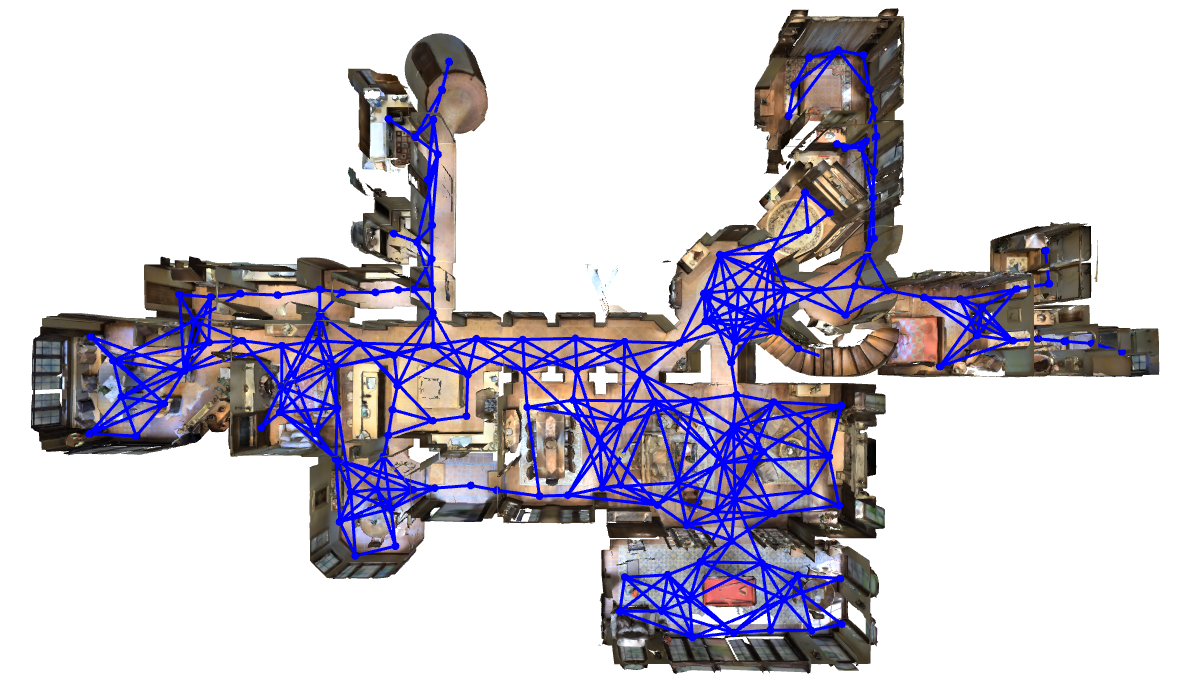}
	\end{center}
	\caption{Example navigation graph for a partial floor of one building-scale scene in the Matterport3D Simulator. Navigable paths between panoramic viewpoints are illustrated in blue. Stairs can also be navigated to move between floors.}
	\label{fig:connectivity}
\end{figure}

\figref{fig:connectivity} illustrates a partial example of a typical navigation graph. On average each graph contains 117 viewpoints, with an average vertex degree of 4.1. This compares favorably with grid-world navigation graphs which, due to walls and obstacles, must have an average degree of less than 4. As such, although agent motion is discretized, this does not constitute a significant limitation in the context of most high-level tasks. Even with a real robot it may not be practical or necessary to continuously re-plan higher-level objectives with every new RGB-D camera view. Indeed, even agents operating in 3D simulators that notionally support continuous motion typically use discretized action spaces in practice~\cite{zhu2017icra, embodiedqa,iqa,savva2017minos}.

The simulator does not define or place restrictions on the agent's goal, reward function, or any additional context (such as natural language navigation instructions). These aspects of the RL environment are task and dataset dependent, for example as described in \secref{sec:dataset}.

\subsubsection{Implementation Details}

The Matterport3D Simulator is written in C++ using OpenGL. In addition to the C++ API, Python bindings are also provided, allowing the simulator to be easily used with deep learning frameworks such as Caffe~\cite{jia2014caffe} and TensorFlow~\cite{abadi2016tensorflow}, or within RL platforms such as ParlAI~\cite{parlai} and OpenAI Gym~\cite{openaigym}. Various configuration options are offered for parameters such as image resolution and field of view. Separate to the simulator, we have also developed a WebGL browser-based visualization library for collecting text annotations of navigation trajectories using Amazon Mechanical Turk, which we will make available to other researchers.

\subsubsection{Biases}

We are reluctant to introduce a new dataset (or simulator, in this case) without at least some attempt to address its limitations and biases~\cite{torralba2011unbiased}. In the Matterport3D dataset we have observed several selection biases. First, the majority of captured living spaces are scrupulously clean and tidy, and often luxurious. Second, the dataset contains very few people and animals, which are a mainstay of many other vision and language datasets~\cite{Chen2015,VQA}. Finally, we observe some capture bias as selected viewpoints generally offer commanding views of the environment (and are therefore not necessarily in the positions in which a robot might find itself). Alleviating these limitations to some extent, the simulator can be extended by collecting additional building scans. Refer to Stanford 2D-3D-S~\cite{2017arXiv170201105A} for a recent example of an academic dataset collected with a Matterport camera.

\section{Room-to-Room (R2R) Navigation}
\label{sec:dataset}

We now describe the Room-to-Room (R2R) task and dataset, including an outline of the data collection process and analysis of the navigation instructions gathered. 

\subsection{Task}

\begin{figure}[t]
	\begin{center}
		\includegraphics[trim={0.7cm 0cm 0.4cm 0},clip,width=1\linewidth]{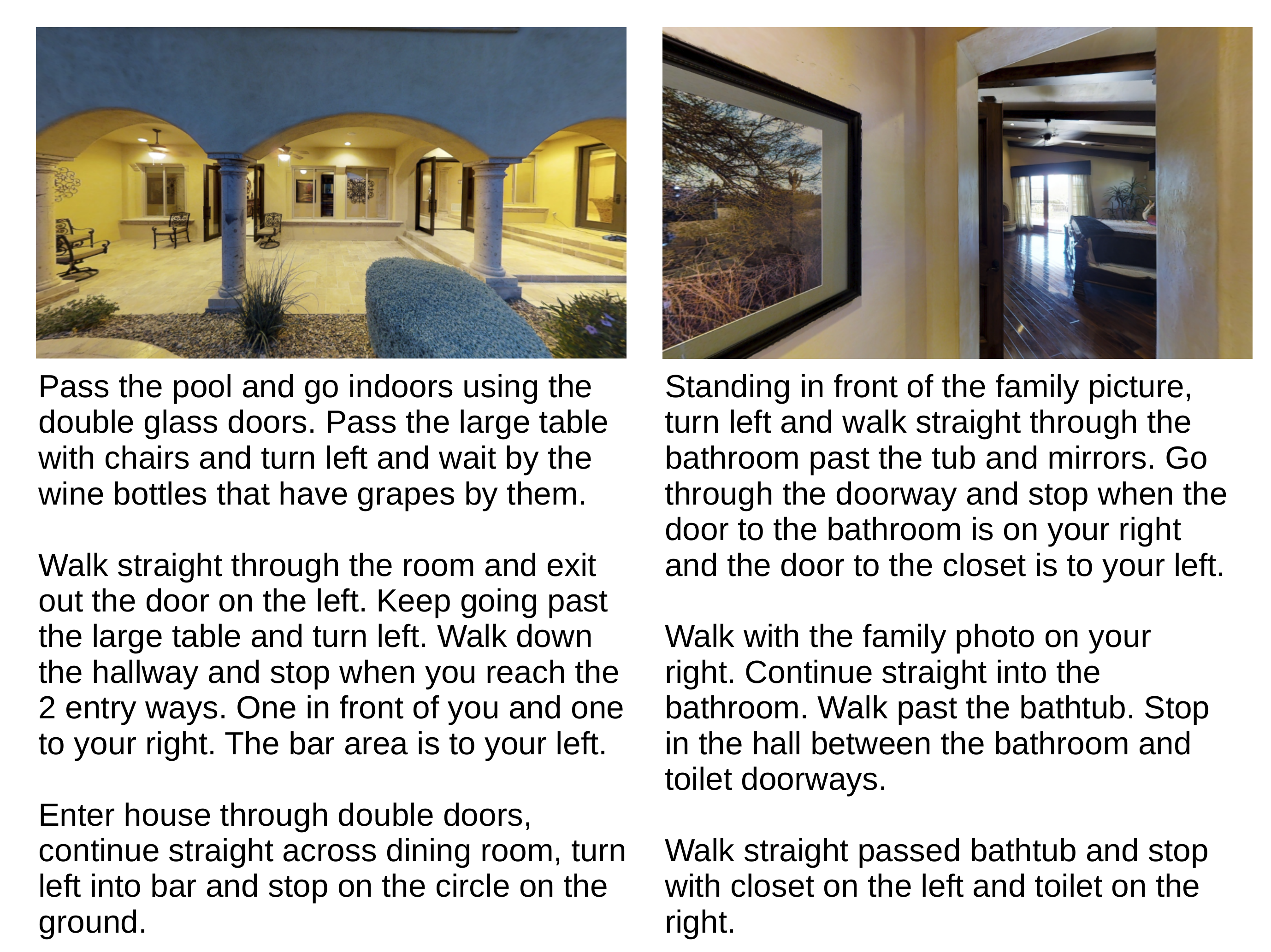}
		\includegraphics[trim={0.7cm 2.7cm 0.4cm 0},clip,width=1\linewidth]{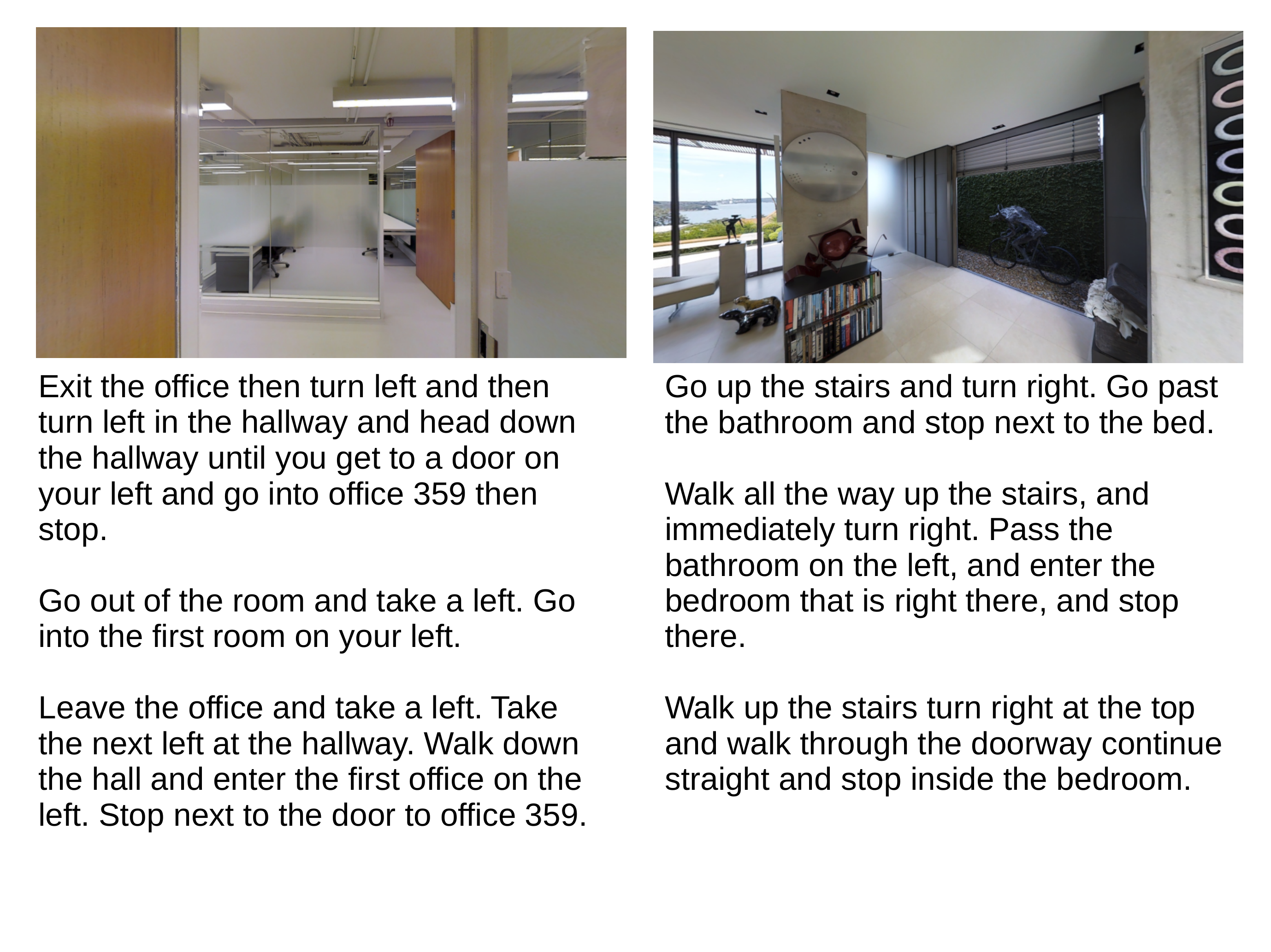}
	\end{center}
	\caption{Randomly selected examples of navigation instructions (three per trajectory) shown with the view from the starting pose.}
	\label{fig:examples}
\end{figure}

As illustrated in \figref{fig:concept}, the R2R task requires an embodied agent to follow natural language instructions to navigate from a starting pose to a goal location in the Matterport3D Simulator. Formally, at the beginning of each episode the agent is given as input a natural language instruction $\bar{x}= \langle x_1,x_2,\dots\,x_L \rangle $, where $L$ is the length of the instruction and $x_i$ is a single word token. The agent observes an initial RGB image $o_0$, determined by the agent's initial pose comprising a tuple of 3D position, heading and elevation $s_0 = \langle v_0, \psi_0, \theta_0 \rangle$. The agent must execute a sequence of actions $ \langle s_0,a_0,s_1,a_1,\dots,s_T,a_T \rangle $, with each action $a_t$ leading to a new pose $s_{t+1} = \langle v_{t+1}, \psi_{t+1}, \theta_{t+1} \rangle$, and generating a new image observation $o_{t+1}$. The episode ends when the agent selects the special \texttt{stop} action, which is augmented to the simulator action space defined in \secref{sec:action-space}. The task is successfully completed if the action sequence delivers the agent close to an intended goal location $v^*$ (refer to \secref{sec:evaluation} for evaluation details).

\subsection{Data Collection}
\label{sec:collection}

\begin{figure}[t]
	\begin{center}
		\includegraphics[trim={0.3cm 0.45cm 0.3cm 0},clip,width=0.49\linewidth]{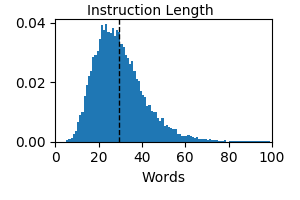}
		\includegraphics[trim={0.3cm 0.45cm 0.3cm 0},clip,width=0.49\linewidth]{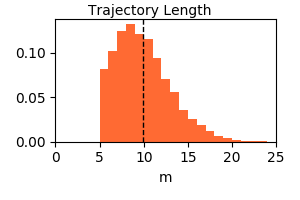}
	\end{center}
	\caption{Distribution of instruction length and navigation trajectory length in the R2R dataset.}
	\label{fig:lengths}
\end{figure}

To generate navigation data, we use the Matterport3D region annotations to sample start pose $s_0$ and goal location $v^*$ pairs that are (predominantly) in different rooms. For each pair, we find the shortest path $v_0:v^*$ in the relevant weighted, undirected navigation graph $G$, discarding paths that are shorter than 5m, and paths that contain less than four or more than six edges. In total we sample 7,189 paths capturing most of the visual diversity in the dataset. The average path length is 10m, as illustrated in \figref{fig:lengths}.

For each path, we collect three associated navigation instructions using Amazon Mechanical Turk (AMT). To this end, we provide workers with an interactive 3D WebGL environment depicting the path from the start location to the goal location using colored markers. Workers can interact with the trajectory as a `fly-through', or pan and tilt the camera at any viewpoint along the path for additional context. We then ask workers to `write directions so that a smart robot can find the goal location after starting from the same start location'. Workers are further instructed that it is not necessary to follow exactly the path indicated, merely to reach the goal. A video demonstration is also provided. 

The full collection interface (which is included as supplementary material) was the result of several rounds of experimentation. We used only US-based AMT workers, screened according to their performance on previous tasks. Over 400 workers participated in the data collection, contributing around 1,600 hours of annotation time.

\subsection{R2R Dataset Analysis}
\label{sec:analysis}

\begin{figure}[t]
	\begin{center}
		\includegraphics[width=0.92\linewidth]{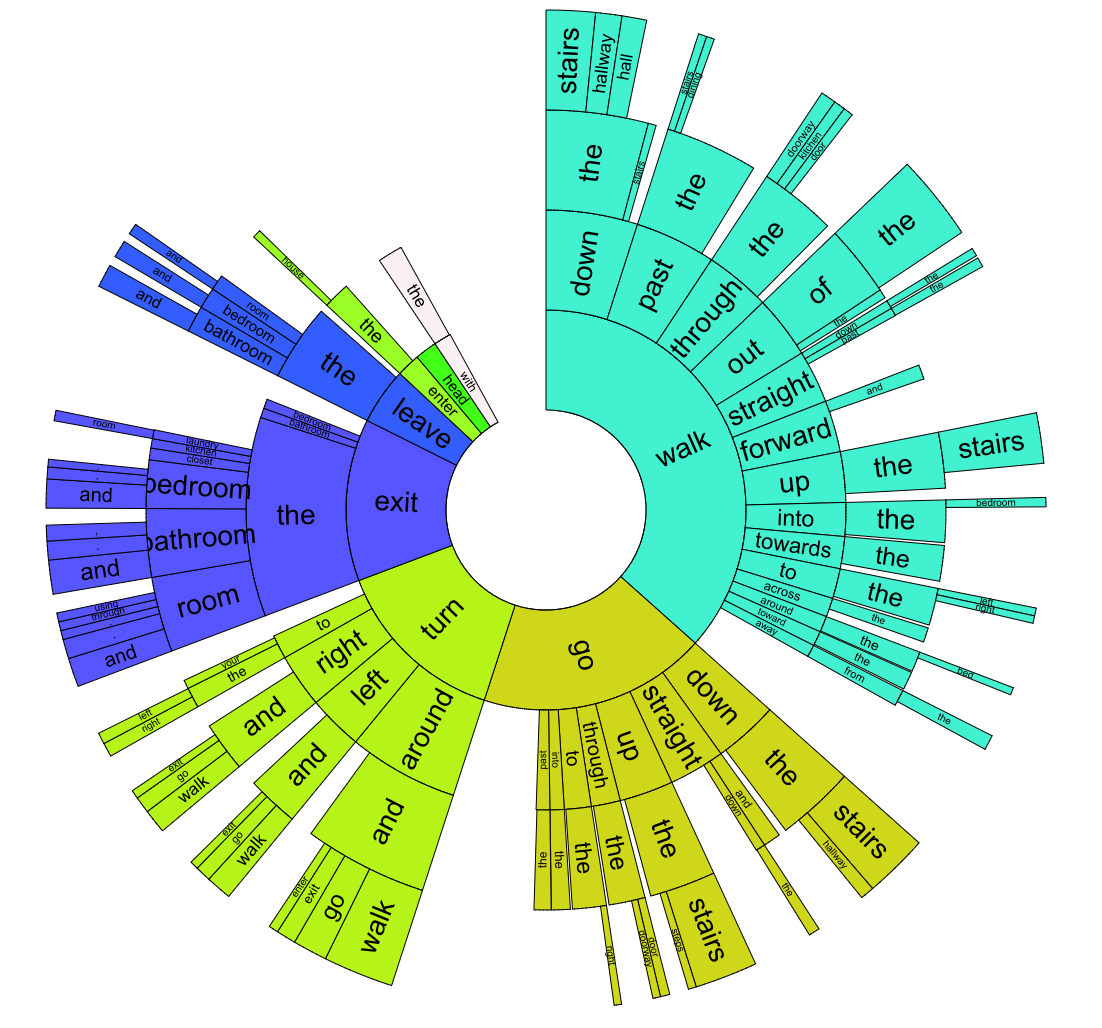}
	\end{center}
	\caption{Distribution of navigation instructions based on their first four words. Instructions are read from the center outwards. Arc lengths are proportional to the number of instructions containing each word. White areas represent words with individual contributions too small to show.}
	\label{fig:distribution}
\end{figure}

In total, we collected 21,567 navigation instructions with an average length of 29 words. This is considerably longer than visual question answering datasets where most questions range from four to ten words~\cite{VQA}. However, given the focused nature of the task, the instruction vocabulary is relatively constrained, consisting of around 3.1k words (approximately 1.2k with five or more mentions). As illustrated by the examples included in \figref{fig:examples}, the level of abstraction in instructions varies widely. This likely reflects differences in people's mental models of the way a `smart robot' works~\cite{Norman:2002}, making the handling of these differences an important aspect of the task. The distribution of navigation instructions based on their first words is depicted in \figref{fig:distribution}. Although we use the R2R dataset in conjunction with the Matterport3D Simulator, we see no technical reason why this dataset couldn't also be used with other simulators based on the Matterport dataset~\cite{Matterport3D}.

\subsection{Evaluation Protocol}
\label{sec:evaluation}

One of the strengths of the R2R task is that, in contrast to many other vision and language tasks such as image captioning and visual dialog, success is clearly measurable. We define \textit{navigation error} as the shortest path distance in the navigation graph $G$ between the agent's final position $v_T$ (i.e., disregarding heading and elevation) and the goal location $v^*$. We consider an episode to be a \textit{success} if the navigation error is less than 3m. This threshold allows for a margin of error of approximately one viewpoint, yet it is comfortably below the minimum starting error of 5m. We do not evaluate the agent's entire trajectory as many instructions do not specify the path that should be taken. 

Central to our evaluation is the requirement for the agent to choose to end the episode when the goal location is identified. We consider stopping to be a fundamental aspect of completing the task, demonstrating understanding, but also freeing the agent to potentially undertake further tasks at the goal. However, we acknowledge that this requirement contrasts with recent works in vision-only navigation that do not train the agent to stop~\cite{zhu2017icra,mirowski2016learning}. To disentangle the problem of recognizing the goal location, we also report success for each agent under an \textit{oracle} stopping rule, i.e. if the agent stopped at the closest point to the goal on its trajectory. Misra \etal~\cite{misra2017mapping} also use this evaluation.

\vspace{-0.3cm}
\paragraph{Dataset Splits}

We follow broadly the same train/val/test split strategy as the Matterport3D dataset~\cite{Matterport3D}. The test set consists of 18 scenes, and 4,173 instructions. We reserve an additional 11 scenes and 2,349 instructions for validating in unseen environments (val unseen). The remaining 61 scenes are pooled together, with instructions split 14,025 train / 1,020 val seen. Following best practice, goal locations for the test set will not be released. Instead, we will provide an evaluation server where agent trajectories may be uploaded for scoring.

\section{Vision-and-Language Navigation Agents}

In this section, we describe a sequence-to-sequence neural network agent and several other baselines that we use to explore the difficulty of the R2R navigation task. 
\subsection{Sequence-to-Sequence Model}

We model the agent with a recurrent neural network policy using an LSTM-based~\cite{Hochreiter1997} sequence-to-sequence architecture with an attention mechanism~\cite{Bahdanau2015}. Recall that the agent begins with a natural language instruction $\bar{x}= \langle x_1,x_2,\dots\,x_L \rangle$, and an initial image observation $o_0$. The encoder computes a representation of $\bar{x}$. At each step $t$, the decoder observes representations of the current image $o_t$ and the previous action $a_{t-1}$ as input, applies an attention mechanism to the hidden states of the language encoder, and predicts a distribution over the next action $a_t$. Using this approach, the decoder maintains an internal memory of the agent's entire preceeding history, which is essential for navigating in a partially observable environment~\cite{wierstra2007solving}. We discuss further details in the following sections. 

\vspace{-0.3cm}
\paragraph{Language instruction encoding} 

Each word $x_i$ in the language instruction is presented sequentially to the encoder LSTM as an embedding vector. We denote the output of the encoder at step $i$ as $h_i$, such that $h_i=\textrm{LSTM}_{enc}\,(x_i,h_{i-1})$. We denote $\bar{h}=\{h_1,h_2,\dots,h_L\}$ as the encoder context, which will be used in the attention mechanism. As with Sutskever \etal~\cite{Sutskever2014}, we found it valuable to reverse the order of words in the input language instruction. 

\vspace{-0.3cm}
\paragraph{Model action space}

The simulator action space is state-dependent (refer \secref{sec:action-space}), allowing agents to make fine-grained choices between different forward trajectories that are presented. However, in this initial work we simplify our model action space to 6 actions corresponding to \texttt{left}, \texttt{right}, \texttt{up}, \texttt{down}, \texttt{forward} and \texttt{stop}. The \texttt{forward} action is defined to always move to the reachable viewpoint that is closest to the centre of the agent's visual field. The \texttt{left}, \texttt{right}, \texttt{up} and \texttt{down} actions are defined to move the camera by 30 degrees.

\vspace{-0.3cm}
\paragraph{Image and action embedding}

For each image observation $o_t$, we use a ResNet-152~\cite{he2015deep} CNN pretrained on ImageNet~\cite{ILSVRC15} to extract a mean-pooled feature vector. Analogously to the embedding of instruction words, an embedding is learned for each action. The encoded image and previous action features are then concatenated together to form a single vector $q_t$. The decoder LSTM operates as $h^{'}_t= \textrm{LSTM}_{dec}\,(q_t,h^{'}_{t-1})$. 

\vspace{-0.3cm}
\paragraph{Action prediction with attention mechanism}

To predict a distribution over actions at step $t$, we first use an attention mechanism to identify the most relevant parts of the navigation instruction. This is achieved by using the global, general alignment function described by Luong \etal~\cite{luong2015effective} to compute an instruction context $c_t = f(h^{'}_t, \bar{h})$. When then compute an attentional hidden state $\tilde{h}_t = \tanh\,(W_c[c_t;h^{'}_t])$, and calculate the predictive distribution over the next action as $a_t = \textrm{softmax}\,(\tilde{h}_t)$. Although visual attention has also proved highly beneficial in vision and language problems~\cite{stacked,coatt,Anderson2017up-down}, we leave an investigation of visual attention in Vision-and-Language Navigation to future work. 

\subsection{Training}

We investigate two training regimes, `teacher-forcing' and `student-forcing'. In both cases, we use cross entropy loss at each step to maximize the likelihood of the ground-truth target action $a_t^*$ given the previous state-action sequence $\langle s_0,a_0,s_1,a_1, \dots, s_t \rangle $. The target output action $a_t^*$ is always defined as the next action in the ground-truth shortest-path trajectory from the agent's current pose $s_t = \langle v_t, \psi_t, \theta_t \rangle$ to the target location $v^*$.

Under the `teacher-forcing'~\cite{lamb2016professor} approach, at each step during training the ground-truth target action $a_t^*$ is selected, to be conditioned on for the prediction of later outputs. However, this limits exploration to only states that are in ground-truth shortest-path trajectory, resulting in a changing input distribution between training and testing~\cite{ross2011reduction,lamb2016professor}. To address this limitation, we also investigate `student-forcing'. In this approach, at each step the next action is sampled from the agent's output probability distribution. Student-forcing is equivalent to an online version of DAGGER~\cite{ross2011reduction}, or the `always sampling' approach in scheduled sampling~\cite{bengio2015scheduled}\footnote{Scheduled sampling has been shown to improve performance on tasks for which it is difficult to exactly determine the best next target output $a_t^*$ for an arbitrary preceding sequence (e.g. language generation~\cite{bengio2015scheduled}). However, in our task we can easily determine the shortest trajectory to the goal location from anywhere, and we found in initial experiments that scheduled sampling performed worse than student-forcing (i.e., always sampling).}.

\vspace{-0.3cm}
\paragraph{Implementation Details}

We perform only minimal text pre-processing, converting all sentences to lower case, tokenizing on white space, and filtering words that do not occur at least five times. We set the simulator image resolution to 640 $\times$ 480 with a vertical field of view of 60 degrees. We set the number of hidden units in each LSTM to 512, the size of the input word embedding to 256, and the size of the input action embedding to 32. Embeddings are learned from random initialization. We use dropout of 0.5 on embeddings, CNN features and within the attention model.

As we have discretized the agent's heading and elevation changes in 30 degree increments, for fast training we extract and pre-cache all CNN feature vectors. We train in PyTorch using the Adam optimizer~\cite{kingma2014adam} with weight decay and a batch size of 100. In all cases we train for a fixed number of iterations. As the evaluation is single-shot, at test time we use greedy decoding~\cite{scst2016}. Our test set submission is trained on all training and validation data.
 
\subsection{Additional Baselines}

\paragraph{Learning free} We report two learning-free baselines which we denote as \textsc{Random} and \textsc{Shortest}. The \textsc{Random} agent exploits the characteristics of the dataset by turning to a randomly selected heading, then completing a total of 5 successful \texttt{forward} actions (when no \texttt{forward} action is available the agent selects \texttt{right}). The \textsc{Shortest} agent always follows the shortest path to the goal.

\vspace{-0.3cm}
\paragraph{Human} We quantify human performance by collecting human-generated trajectories for one third of the test set (1,390 instructions) using AMT. The collection procedure is similar to the dataset collection procedure described in \secref{sec:collection}, with two major differences. First, workers are provided with navigation instructions. Second, the entire scene environment is freely navigable in first-person by clicking on nearby viewpoints. In effect, workers are provided with the same information received by an agent in the simulator. To ensure a high standard, we paid workers bonuses for stopping within 3m of the true goal location.

\begin{figure*}[t]
	\begin{center}
		\includegraphics[width=1\textwidth]{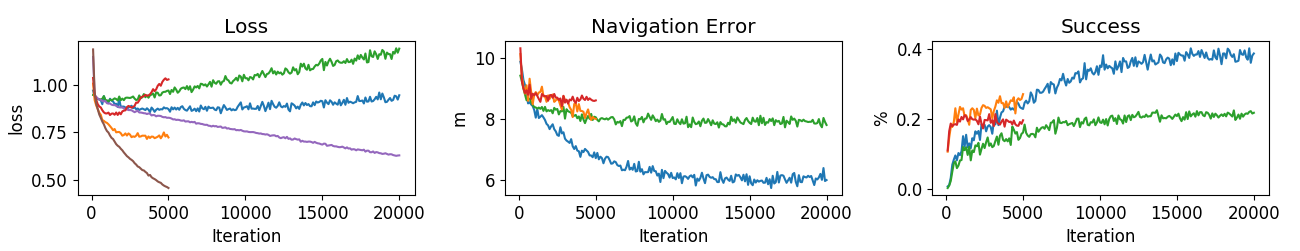}
		\includegraphics[trim={0cm 0.2cm 0cm 0.2cm},clip,width=1\textwidth]{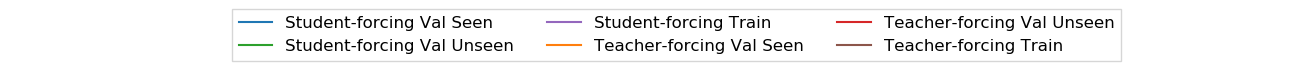}
	\end{center}
	\caption{Validation loss, navigation error and success rate during training. Our experiments suggest that neural network approaches can strongly overfit to training environments, even with regularization. This makes generalizing to unseen environments challenging. }
	\label{fig:training}
\end{figure*}

\setlength{\tabcolsep}{.45em}
\begin{table}[]
	\footnotesize
	\begin{center}
		\begin{tabular}{lcccc}
			\midrule
			& \makecell{\textbf{Trajectory}\\ \textbf{Length} \textbf{(m)}} & \makecell{\textbf{Navigation}\\ \textbf{Error} \textbf{(m)}} & \makecell{\textbf{Success}\\ \textbf{(\%)}} & \makecell{\textbf{Oracle}\\ \textbf{Success} \textbf{(\%)}} \\
			\midrule
			\textbf{Val Seen:}   &                           &                          &                  &                  \\
			\textsc{Shortest}      &           10.19                &    0.00                      &      100           &        100          \\
			\textsc{Random}      &           9.58               &    9.45                      &       15.9           &        21.4          \\
			Teacher-forcing      &         10.95                  &       8.01                   &       27.1           &         36.7        \\
			Student-forcing      &         11.33                  &      6.01                    &        38.6          &         52.9        \\
			\midrule
			\textbf{Val Unseen:} &                           &                          &                  &                  \\
			\textsc{Shortest}      &         9.48                  &       0.00                   &     100             &       100           \\
			\textsc{Random}      &          9.77                 &      9.23                    &      16.3            &          22.0        \\
			Teacher-forcing      &            10.67               &        8.61                  &      19.6            &       29.1           \\
			Student-forcing      &       8.39                    &       7.81                   &      21.8           &      28.4            \\
			\midrule     
			\textbf{Test (unseen):} &                           &                          &                  &                  \\
			\textsc{Shortest}      &         9.93                  &       0.00                   &     100             &       100           \\
			\textsc{Random}      &         9.93                  &      9.77                    &      13.2            &      18.3            \\
			Human      &          11.90                 &            1.61              &       86.4           &        90.2         \\
			Student-forcing            &         8.13                 &      7.85                   &  20.4   &         26.6         \\       
			\midrule
		\end{tabular}
	\end{center}
	\caption{Average R2R navigation results using evaluation metrics defined in \secref{sec:evaluation}. Our seq-2-seq model trained with student-forcing achieves promising results in previously explored environments (Val Seen). Generalization to previously \textit{unseen} environments (Val Unseen / Test) is far more challenging. }
	\label{tab:results}
\end{table}

\begin{figure}[t]
	\begin{center}
		\includegraphics[trim={0cm 0.4cm 0cm 0.4cm},clip,width=1\linewidth]{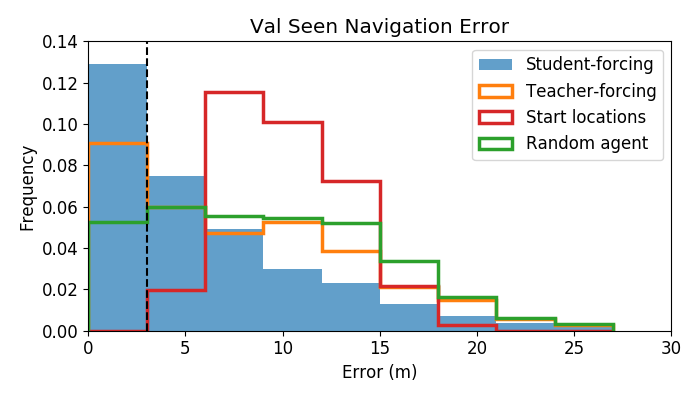}
	\end{center}
	\caption{In previously seen environments student-forcing training achieves 38.6\% success ($<$ 3m navigation error). }
	\label{fig:nav-error}
\end{figure}

\section{Results}
\label{sec:results}

As illustrated in \tabref{tab:results}, our exploitative \textsc{Random} agent achieves an average success rate of 13.2\% on the test set (which appears to be slightly more challenging than the validation sets). In comparison, AMT workers achieve 86.4\% success on the test set, illustrating the high quality of the dataset instructions. Nevertheless, people are not infallible when it comes to navigation. For example, in the dataset we occasionally observe some confusion between right and left (although this is recoverable if the instructions contain enough visually-grounded references). In practice, people also use two additional mechanisms to reduce ambiguity that are not available here, namely gestures and dialog.

With regard to the sequence-to-sequence model, student-forcing is a more effective training regime than teacher-forcing, although it takes longer to train as it explores more of the environment. Both methods improve significantly over the \textsc{Random} baseline, as illustrated in \figref{fig:nav-error}. Using the student-forcing approach we establish the first test set leaderboard result achieving a 20.4\% success rate.

The most surprising aspect of the results is the significant difference between performance in seen and unseen validation environments (38.6\% vs. 21.8\% success for student-forcing). To better explain these results, in \figref{fig:training} we plot validation performance during training. Even using strong regularization (dropout and weight decay), performance in unseen environments plateaus quickly, but further training continues to improve performance in the training environments. This suggests that the visual groundings learned may be quite specific to the training environments. 

Overall, the results illustrate the significant challenges involved in training agents that can generalize to perform well in previously unseen environments. The techniques and practices used to optimize performance on existing vision and language datasets are unlikely to be sufficient for models that are expected to operate in new environments.

\section{Conclusion and Future Work}

Vision-and-Language Navigation (VLN) is important because it represents a significant step towards capabilities critical for practical robotics. To further the investigation of VLN, in this paper we introduced the Matterport3D Simulator. This simulator achieves a unique and desirable trade-off between reproducibility, interactivity, and visual realism. Leveraging these advantages, we collected the Room-to-Room (R2R) dataset. The R2R dataset is the first dataset to evaluate the capability to follow natural language navigation instructions in previously unseen real images at building scale. To explore this task we investigated several baselines and a sequence-to-sequence neural network agent.

From this work we reach three main conclusions. First, VLN is interesting because existing vision and language methods can be successfully applied. Second, the challenge of generalizing to previously \textit{unseen} environments is significant. Third, crowd-sourced reconstructions of \textit{real} locations are a highly-scalable and underutilized resource\footnote{The existing Matterport3D data release constitutes just 90 out of more than 700,000 building scans that have been already been collected~\cite{Matt}.}. The process used to generate R2R is applicable to a host of related vision and language problems, particularly in robotics. We hope that this simulator will benefit the community by providing a visually-realistic framework to investigate VLN and related problems such as navigation instruction generation, embodied visual question answering, human-robot dialog, and domain transfer to real settings.

{
\small
\textbf{Acknowledgements} This research is supported by a Facebook ParlAI Research Award, an Australian Government Research Training Program (RTP) Scholarship, the Australian Research Council Centre of Excellence for Robotic Vision (project number CE140100016), and the Australian Research Council's Discovery Projects funding scheme (project DP160102156).
}

{\small
\bibliographystyle{ieee}
\bibliography{paper}
}

\newpage

\begin{figure*}[h]
	\begin{center}
		\includegraphics[width=1\linewidth]{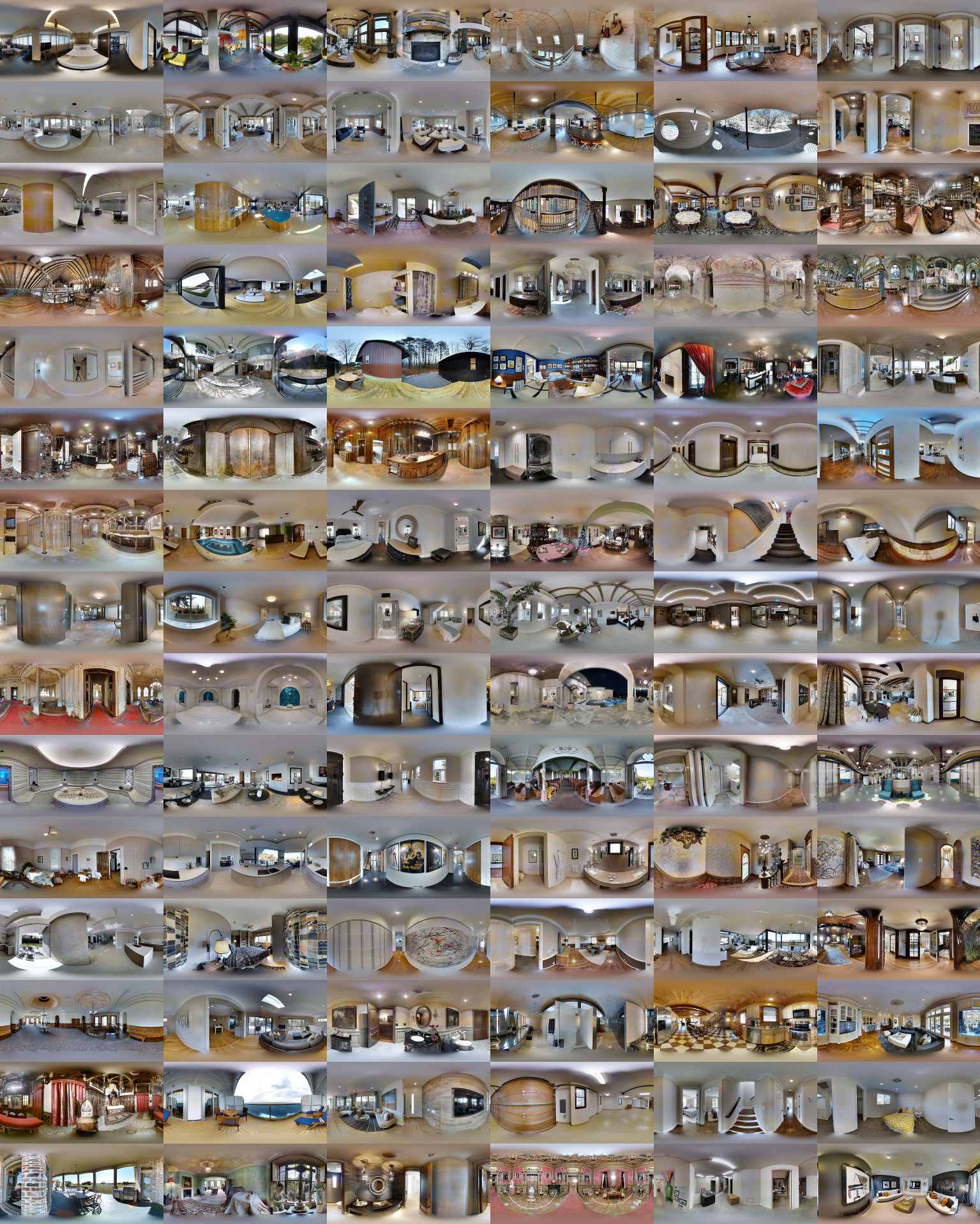}
	\end{center}
	\caption{Snapshot of the visual diversity in the Matterport3D dataset, illustrating one randomly selected panoramic viewpoint per scene. }
	\label{fig:tiles}
\end{figure*}

\begin{figure*}
	\begin{center}
		\makebox[\textwidth][c]{\includegraphics[width=1.0\linewidth]{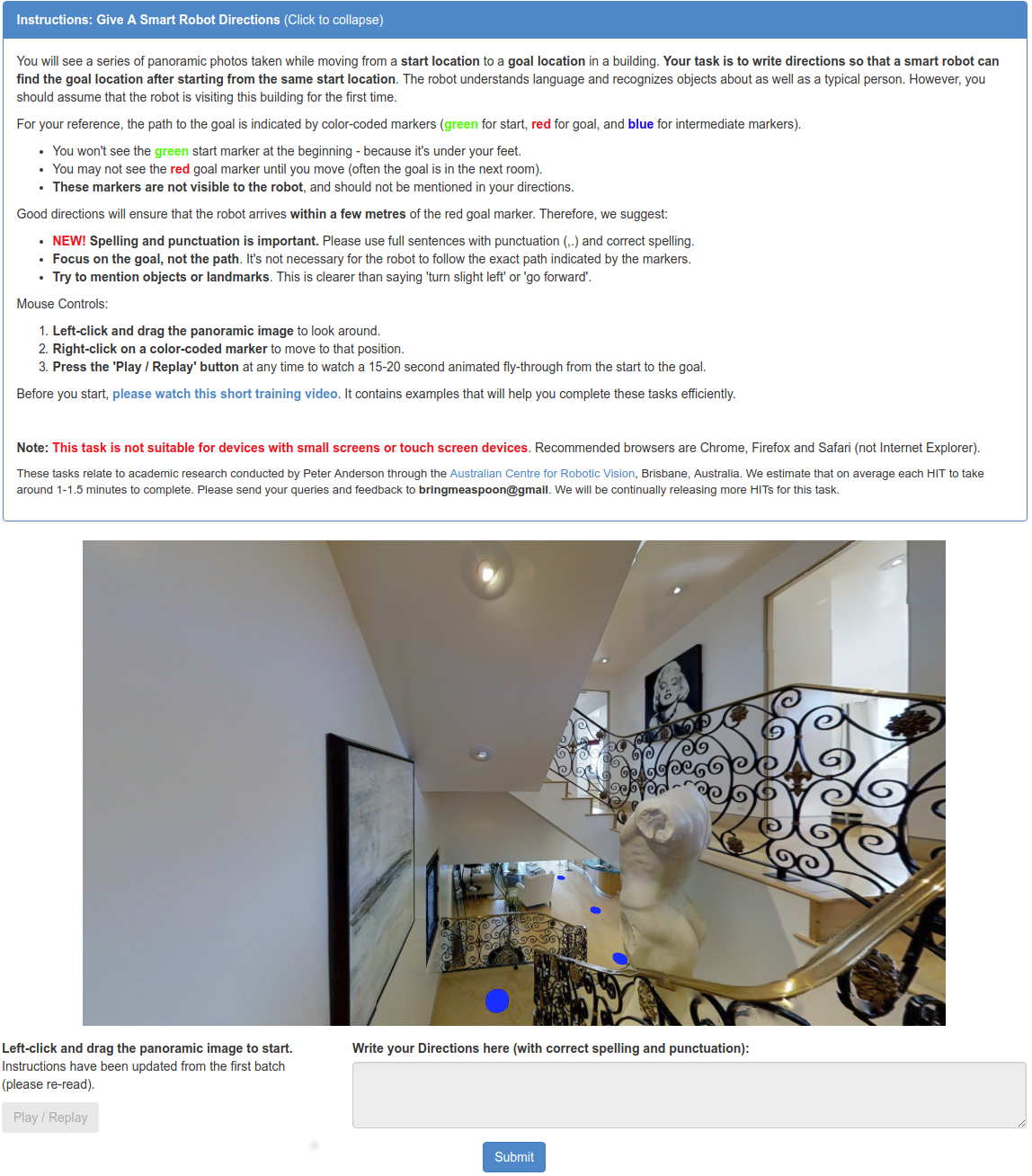}}
	\end{center}
	\caption{AMT data collection interface for the R2R navigation dataset. Here, blue markers can be seen indicating the trajectory to the goal location. However, in many cases the worker must first look around (pan and tilt) to find the markers. Clicking on a marker moves the camera to that location. Workers can also watch a `fly-through' of the complete trajectory by clicking the Play / Replay button. }
	\label{fig:interface}
\end{figure*}

\begin{table*}[t]
	\small
	\begin{center}
		\begin{tabular}{p{17cm}}
			\toprule
			Go past the ovens and the counter and wait just before you go outside. \\
			Walk through the kitchen towards the living room. Walk around the island and step onto the patio near the two chairs and stop in the patio doorway. \\
			Exit the kitchen by walking past the ovens and then head right, stopping just at the doorway leading to the patio outside. \\
			\midrule
			Go up the last few stairs and turn right. Go up the next two flights of stairs and wait. \\
			Walk up the rest of the stairs, then continue up the next set of stairs.  Stop at the top of the stairs near the potted plant. \\
			Go up the stairs then turn right and go up the other stairs on the right then turn right and go up the other stairs on the right and stop at the top of the stairs. \\
			\midrule
			Walk until your in the next room. Make a right into the room on the right. Stop in front of the water heater. \\
			Go across the room opposite the brown door, make a sharp right turn, and take a step into the laundry room and stop. \\
			Exit the room. Turn right and then right again into the room next door. Wait there. \\
			\midrule
			Turn right and enter the bedroom. Cross the bedroom and turn right and stop at the door leading out of the bedroom. \\
			From shower room enter bedroom, walk across bedroom to hall and stop at window. \\
			Exit the bathroom toward the bedroom. Exit the bedroom using the door on the right. \\
			\midrule
			Turn around a the blackboard, make a left at the water fountain and head through the doorframe. Angle left and move straight, keeping the table with the white tablecloth on your left side. Make a slight right and walk straight, waiting at the bottom of the stairwell. \\
			Walk towards the water dispenser and exit the doorway to the left. Walk straight left of the white circular table and towards the wooden staircase. \\
			Walk out of the bathroom, turn left, and wait at the bottom of the stairs. \\
			\midrule
			Walk toward the bed. When you get to the bed. Turn right and exit the room. Continue straight and enter the room straight ahead. Wait near the sink. \\
			Turn to the left and enter the bedroom. Once inside, turn right and walk straight ahead and stop when you enter the bathroom. \\
			Exit the bathroom, then turn left. Wait in the office next to the desk. \\
			\midrule
			Turn around and go up the stairs, turn right and go to right again towards the front door. \\
			Make your way up to the steps and then pull a hard right followed by another hard right after three steps. then continue until you've reached the first open door and stop. \\
			Walk up the stairs and turn hard right. Stop in the bathroom doorway on the left. \\
			\midrule
			Turn and enter the living room area. Go past the table and sofas and stop in the foyer in front of the front door. \\
			Turn around and exit the room. Walk around the sofa and enter the hallway. Wait by the side table. \\
			Exit the room through the doorway nearest you, and continue into the adjacent room, exiting the room via the exit to your left. \\
			\midrule
			Turn right towards kitchen. Go into hallway and walk into dining room. \\
			walk through the archway with the thermostat on the wall.  Walk toward the piano and stop just before it. \\
			Turn toward the kitchen, and walk through the doorway to the right of the breakfast bar.  Walk down the hall, passing the bathroom on your right side as you walk.  Walk straight and stop when you get to the piano. \\
			\midrule
			Walk along the insulated bare walls towards the window ahead in the next room. Walk through the unfinished room and through the door on the other side of the room that leads to a finished hallway. Walk into the first open door in the hall that leads to a bedroom with photo art on the wall near the entrance of classic black and white scenes. \\
			Walk forward past the window then turn right and enter the hallway.  Enter the first bedroom on your right. wait near the bed. \\
			Walk forward and take a right. Enter the hallway through the door on the right. Take the first left into a bedroom. Stop once you are in the bedroom. \\
			\bottomrule
		\end{tabular}
	\end{center}
	\caption{Examples of randomly selected R2R navigation instructions. Each cell contains three instructions associated with the same path. }
	\label{tab:instructions}
\end{table*}

\end{document}